\documentclass{article}
\usepackage{spconf,amsmath,graphicx}
\usepackage{subcaption}
\usepackage{caption}
\usepackage{graphicx}
\usepackage{array}
\usepackage{amsmath}
\usepackage{nicefrac}
\usepackage{amsfonts}
\usepackage{amssymb}
\usepackage{amsthm}
\title{A MULTI-PERSPECTIVE APPROACH TO ANOMALY DETECTION FOR SELF-AWARE EMBODIED AGENTS}
\name{{\footnotesize Mohamad Baydoun$^{1,2}$,Mahdyar Ravanbakhsh$^{1}$,Damian Campo$^{1}$,Pablo Marin$^{3}$, David Martin$^{3}$, Lucio Marcenaro$^{1}$, Andrea Cavallaro$^{2}$, Carlo S. Regazzoni$^{1,3}$}}
	
\address{
\small
$^1$DITEN, University of Genoa, Italy.
\small
$^2$CIS, Queen Mary University of London, UK.\\
\small
$^3$Carlos III University of Madrid, Spain.\\
}
\begin{document}
\ninept
\maketitle
\begin{abstract}
This paper focuses on multi-sensor anomaly detection for moving cognitive agents using both external and private first-person visual observations. Both observation types are used to characterize agents’ motion in a given environment. The proposed method generates locally uniform motion models by dividing a Gaussian process that approximates agents’ displacements on the scene and provides a Shared Level (SL) self-awareness based on Environment Centered (EC) models. Such models are then used to train in a semi-unsupervised way a set of Generative Adversarial Networks (GANs) that produce an estimation of external and internal parameters of moving agents. Obtained results exemplify the feasibility of using multi-perspective data for predicting and analyzing trajectory information.
\end{abstract}
\section{Introduction}
\label{sec:intro}
Fully autonomous systems need perception to navigate through scenes and recognize objects in real environments \cite{Ramík2014}. Recent advances in signal processing and machine learning techniques can be useful to design autonomous systems equipped with a self-awareness module that facilitates to recognize contextual information while a given task is executed. The capability of detecting abnormal situations based on such self-awareness is an important task that allows autonomous systems to increase their situational awareness and the effectiveness of the decision making submodules \cite{Campo2017}.
 
The analysis of observed moving agents for understanding normal/abnormal dynamics in a given scene represents an emerging research field \cite{Bastani2016,Campo2017,Morris2008}. This paper proposes a methodology for abnormality detection based on multiple sensors that observe the same phenomenon from different perspectives. Abnormalities can be first detected as deviations from Environment Centered (EC) models, i.e.,  from an observer viewpoint which does not have access to internal agent variables. Such layer can be defined as a Shared Level (SL) of self-awareness, since the observed information, e.g., observed position and velocity can be measured easily from an external observer.

An observed agent can also have further information corresponding to what it can observe from a first person viewpoint (FPV) while a task is performed. Abnormalities related to unexpected observations acquired while performing a task can be considered as the essential information to define a Private Layer (PL) of self-awareness: Such experiences are available only to the agent itself. Accordingly, an external observer cannot access to such information and has to rely solely on SL information. 

Analyzing phenomena from different sensory data is definitively not a new problem. In \cite{Moll2010}, researchers use video data together with orientation information to capture the 3D motion of a human body. In \cite{Charlon2013}, a multi-sensor monitoring system is proposed to prevent accidents and detect falls. Additionally, several researchers used multiple-cameras to recognize abnormalities \cite{Ermis2008,Emonet2011}. %These works have motivated the use multiple perspectives in order to characterize agents' dynamics, perform predictions and detect unseen behaviors (abnormalities) in environments. 

One of the main novelties of this work consists of a strategy which processes data from first and external viewpoints and facilitates a subdivision of subject's behaviors into basic dynamical models (activities). Dynamic normality models and related algorithms can detect abnormalities by fusing shared and private agent information. Identified anomalies are here defined as patterns that have not seen or learned in previous experiences \cite{Kim2011,Bastani2016} by taking into consideration private and shared perspectives of the same phenomenon.        

%Phenomena here analyzed consist in events where spatial actions taken by an agent, that can be estimated starting from environment centered trajectories, can be enriched by private video sequences the agent can collect while performing such actions.
Observations acquired from the external viewpoint (EV) are composed of agents' positions and velocity with respect a fixed reference system. Locations and velocities (actions) are analyzed by a Gaussian Process (GP) regression that estimates the most probable agent's action in each position of the whole scene. Results from the GP regression are then clustered in zones through a superpixel algorithm approach introduced in \cite{Campo2017b} that encodes action patterns, e.g., going straight or curving. %In this way, a map where simple dynamical models that encode action patterns in the environment can be obtained and used for further processes and forming SL of self awareness. 

Images collected synchronously from an FPV are used to compute optical flow at each frame. Video sequences can be seen as PL data related to the reference system of the moving agent itself.  The optical flow between video frames can be seen as a private representation of the agent's action in the PL. Accordingly, in order to understand the relation between a given FPV image and its optical flow, a Generative Adversarial Nets (GANs) approach \cite{NIPS2014_5423} is adopted for training deep networks in a supervised way. Supervision here consists in supposing that the action patterns previously obtained by the GP approach for defining EC normality can be used to train normal visual models.%, i.e. expected normal dynamic evolution of video sequences in each region exhibiting a uniform EC motion.

Both SL and PL learned models can be used to predict the dynamics of a vehicle performing a task. Produced models by each approach can be used to describe how these two representations are related to each other in normal conditions. This paper shows that the PL representation learned in a supervised way can provide further normality descriptors, enriching the ones obtained in an unsupervised way from GP for EC normality representation.

The remainder of the papers is organized as follows: section \ref{sec:Method} describes the GP approach (section \ref{ssec:RDM}) and the GAN (section \ref{ssec:lgan}) for SL and PL self awareness, respectively. Section \ref{sec:Dataset} shows the dataset that was used for obtaining results that are detailed in sections \ref{ssec:FMR} and \ref{ssec:SAR}. Conclusions and future research directions are presented in section \ref{sec:Conclusions}.

%Results suggest the capability of our methodology at recognizing abnormalities based on a multiple viewpoints. Obtained outcomes of the proposed approach suggest the possibility of future research oriented to the use combined perspectives for decision making where awareness of situations and self-reactions is increased with respect other methods in literature. 
%
%The paper can be divided as follows: Section \ref{sec:Method} explains the current method for recognition of abnormalities from different viewpoints. Section \ref{sec:Dataset} explains the dataset used for testing the proposed algorithms. Obtained results are described in Section \ref{sec:Results}. Some conclusions and future work are discussed in Section \ref{sec:Conclusions} 

\section{Proposed Method}\label{sec:Method}

\subsection{Representation of observed dynamic motion}\label{ssec:RDM}
To model the SL self-awareness, we use a state space representation from an external observer placed in the EC reference system. Accordingly, the state subspace $X$ represents the location of an agent in the environment whereas $\dot{X}$ represents its velocity. the whole scene can be seen as a grid of possible locations where agents can be. Let such spatial grid be defined as $\tilde{X} = \{{X}^1,{X}^2,...,\dot{X}^M \}$. As can be seen, $\tilde{X}$ is a set of $M$ locations that cover the whole environment. Given a set of observations from a moving agent, it is possible to express its positions and velocities as the subsets $X^{*} = \{{X}_1,{X}_2,...,\dot{X}_K \}$ and $\dot{X}^{*} = \{{X}_1,{X}_2,...,\dot{X}_K \}$, where $K$ is the total number of observations of the agent. By using $X^{*}$ and $\dot{X}^{*}$, it is possible to use a GP approach \cite{Kim2011} that approximates velocity information over spatial grid, $\tilde{X}$, such that:
\begin{equation}\label{eq1}
 \tilde{\dot{X}} = g(\tilde{X}) + v,
\end{equation}
where $\tilde{\dot{X}}$ represents an estimation of velocity information for each point of the spatial grid, $g(\cdot)$ takes location information and estimates the expected motion (action) at such position for a given activity. Since agents' actions can be seen as velocities, it is possible to describe them as a GP, where $v$ is a Gaussian zero-mean white noise.

In this work, a 2-dimensional case is considered. Therefore, spatial coordinates and time derivatives consist of two components each, $(x,y)$ and $(\dot{x},\dot{y})$, respectively. Accordingly, it is possible to represent $(x,y)$ as the pixels of an image whose corresponding color values carry information about agents' actions $(\dot{x},\dot{y})$. In particular, here RGB images are considered, where Red and Blue colors encode $(\dot{x},\dot{y})$ respectively and the Green channel is disregarded.

\begin{figure}[ht] 
	\centering
	\includegraphics[width=1\linewidth]{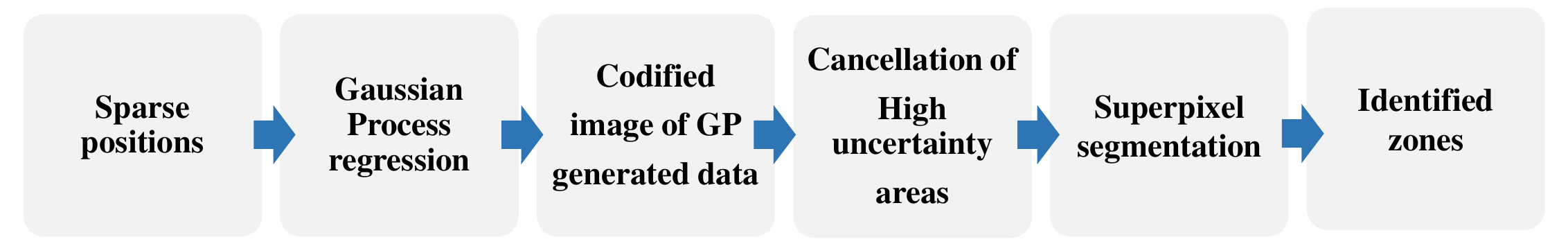} 
	\caption{Identification of dynamic motion in terms of zones} 
	\label{fig00}
	%\vspace{2ex} 
\end{figure}

Uncertainties generated by the GP are used to remove information where not enough evidence is observed such as explained in \cite{Campo2017b}. Since an image that encodes the GP is available, a superpixel algorithm \cite{Zhengqin2015} is applied to discretize the image plane into $N$ zones. Each of these zones is characterized by a quasi-constant velocity model $[\dot{x}_n$, $\dot{y}_n]$, where $n$ indexes identified zones, i.e., $n \in \{1,2,\dots,N \}$.  Finally, a linear dynamic model can be defined for each zone such as follows: 
\begin{equation} \label{eq2}
A_n := X_{k+1}= X_{k} + \Delta k U_{n,k} + w_{m}, 
\end{equation}
where $U_{n,k} = [\dot{x}_n$, $\dot{y}_n]^T$, $k$ indexes the time, $\Delta k$ is the sampling time and $w_{m}$ is the process noise. The variable $U_{n,k}$ is a control input that encodes the action (motivation) of the agent. The process for identifying zones where quasilinear models are valid is summarized in the block diagram in Fig.\ref{fig00}.

\subsubsection{Abnormality detection by using Kalman filter method} \label{ssec:AD}
It is possible to build a set of Kalman Filters (KFs) based on the built dynamical models shown in equation \eqref{eq2}. Each KF is designed for tracking linear motions with low error (innovation) when observed data follows already characterized (normal) behaviors inside identified zones. 

As is well known, KFs' innovations represent residual values produced by measurements while assuming a specific normal model. Such values can be used to express abnormalities since they quantify the deviations from normal learned models in the environment. Innovations can be expressed as: 
\begin{equation}\label{eq01}
\epsilon_{k,n} = Z_k - \hat{X}^n_{k|k-1},
\end{equation}

where $\epsilon_{k,n}$ is the innovation generated in the zone $n$ where the agent is located. $Z_k$ represents observed spatial data and $\hat{X}^n_{k|k-1}$ is the KF estimation of the agent's location at the future time $k$ calculated in the time instant $k-1$  \eqref{eq2}.

In this work, innovation vectors are composed of two components (one for each axis) and the magnitude of those vectors can be considered as a final measure of abnormality, $\xi$, assuming that the observed agent is inside the region $n$, it is possible write:
$$\xi_k = || \epsilon_{k,n} ||_2,$$
In order to evaluate if an observation is abnormal with respect to the current bank of KFs that encodes the normality in an environment (eq. \eqref{eq2}), an error threshold $\xi_{thres}$ is defined for distinguish between abnormal and normal behaviors at each time instant. Accordingly, if a certain $\xi_k$ exceeds such threshold, the system considers the current measurement $Z_k$ as an anomaly. 
\subsection{Representation of the agent embodied self awareness}
In order to represent the PL of agent self-awareness, Generative Adversarial Networks (GANs) \cite{NIPS2014_5423} are proposed to learn the normality pattern of the observed scene. GANs are deep networks commonly used to generate data (e.g., images) and are trained using only unsupervised data. The supervisory information in a GAN is indirectly provided by an adversarial game between two independent networks: a generator ($G$) and a discriminator ($D$). 
During training, $G$ generates new data and $D$ tries to understand whether its input is real  (i.e., it is a training image) or produced by $G$. The competition between $G$ and $D$ is helpful for boosting the ability of both $G$ and $D$.

\subsubsection{Learning the normal pattern of the observed scene}
\label{ssec:lgan}
Two channels are used to learn the normal pattern of the observed scene: appearance (i.e., raw-pixels) and motion (optical flow images) for two cross-channel tasks. In the first task, optical-flow images are generated from the original frames. In the second task, appearance information is estimated from an optical flow image.
\begin{figure}[t] 
	\centering
	\includegraphics[width=1\linewidth]{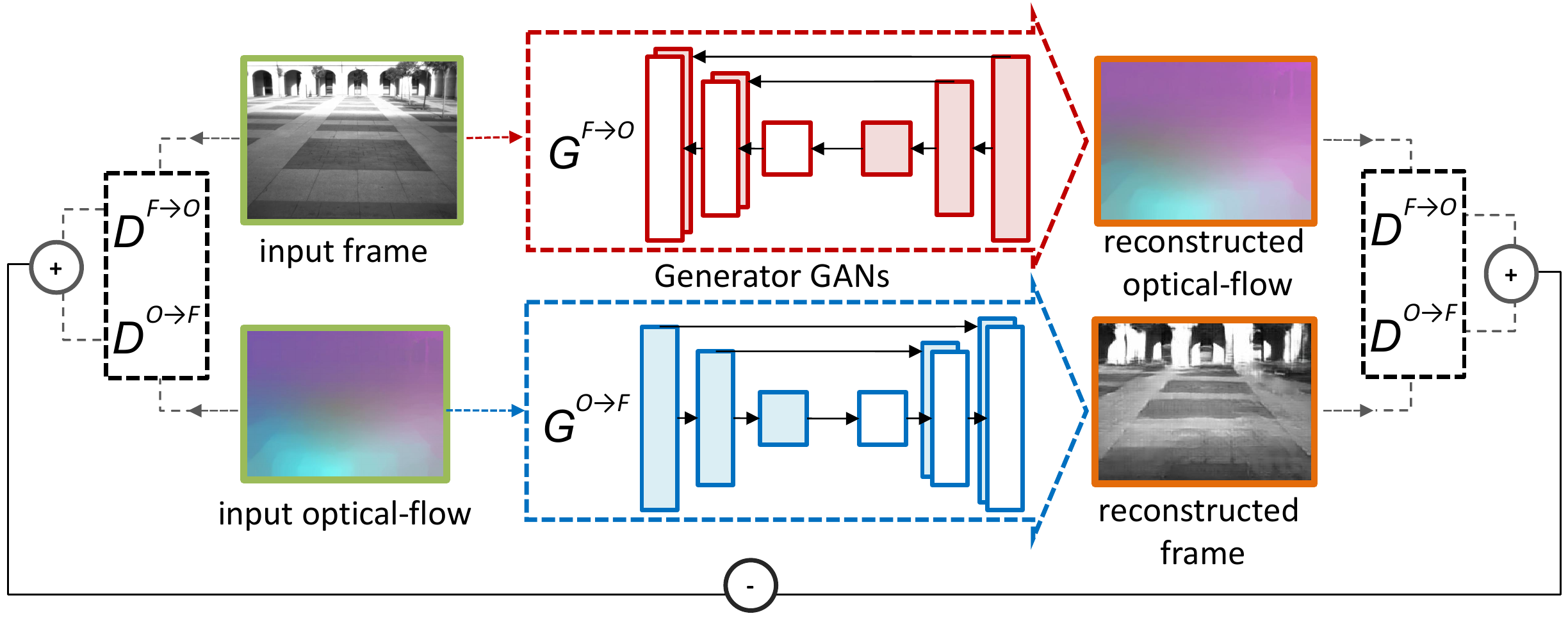}
	\caption{The two GANs structure
		%Two GANs: (i) ${\cal N}^{F \rightarrow O}$ generates optical-flow from frames by ${G}^{F \rightarrow O}$, and (ii) ${\cal N}^{O \rightarrow F}$generates frames from optical-flow by ${G}^{O \rightarrow F}$. Following with the corresponding Discriminators ${D}^{F \rightarrow O}$, and ${G}^{O \rightarrow F}$
	} 
	\label{fig:gan_overview} 
\end{figure}

Let $F_t$ be the $t$-th frame of a training video and $O_t$ the optical flow obtained using $F_t$ and $F_{t+1}$. $O_t$ is computed using \cite{brox2004high}. Fig.\ref{fig:gan_overview} shows two networks: ${\cal N}^{F \rightarrow O}$, which is trained to generate optical-flow from frames (task 1) and ${\cal N}^{O \rightarrow F}$, which generates frames from optical-flow (task 2).
In both cases, inspired by \cite{ravanbakhsh2017training,icip17}, our networks are composed of a conditional generator $G$ and a conditional discriminator $D$. $G$ takes as input an image $x$ and a noise vector $z$ (drawn from a noise distribution ${\cal Z}$) and outputs an image $r = G(x,z)$ of the same dimensions of $x$ but represented in a different channel. 
%For instance, in case of ${\cal N}^{F \rightarrow O}$, $x$ is a frame ($x  = F_t$) and $r$ is {\em the reconstruction} of its corresponding optical-flow image $y  = O_t$. On the other hand, $D$ takes as input two images $x$ and $u$ (where $u$ is either $y$ or $r$) and outputs a scalar representing the probability that both its input images came from the real data.

Both $G$ and $D$ are fully-convolutional networks, composed of convolutional, batch-normalization layers and ReLU nonlinearities. In case of $G$, we adopt the U-Net architecture \cite{DBLP:journals/corr/IsolaZZE16}, which is an encoder-decoder. %where the input $x$ is passed through a series of progressively downsampling layers until a bottleneck layer, at which point the forwarded information is upsampled. Downsampling and upsampling layers in a symmetric position with respect to the bottleneck layer are connected by {\em skip connections} which help preserving important local information.
$D$ is proposed to be a {\em PatchGAN} discriminator \cite{DBLP:journals/corr/IsolaZZE16}, which is based on a ``small'' fully-convolutional discriminator $\hat{D}$.
%
%$G$ and $D$ are trained using both a conditional GAN loss ${\cal L}_{cGAN}$ and a reconstruction loss ${\cal L}_{L1}$. In case of ${\cal N}^{F \rightarrow O}$, the training set is composed of pairs of frame-optical flow images
%${\cal X} = \{ (F_t, O_t) \}_{t=1,...,N}$. 
%${\cal L}_{L1}$ is given by:
%
%\begin{equation}
%{\cal L}_{L1}(x,y) =  ||y - G(x,z) ||_1,
%\end{equation}
%
%\noindent
%where $x = F_t$ and $y = O_t$, whereas the conditional adversarial loss ${\cal L}_{cGAN}$ is:
%
%\begin{equation}
%\begin{split}
%{\cal L}_{cGAN}(G,D)= 
%\mathbb{E}_{(x,y) \in {\cal X}} [\log D(x,y)] +\\
%\mathbb{E}_{x \in \{ F_t \},z \in {\cal Z}} [\log ( 1 - D(x,G(x,z)) )]
%\end{split}
%\end{equation}
%
%In case of ${\cal N}^{O \rightarrow F}$, we use ${\cal X} = \{ (O_t, F_t) \}_{t=1,...,N}$. 
Additional details about the training procedure can be found in \cite{DBLP:journals/corr/IsolaZZE16,icip17}. 
During training, the output of $\hat{D}$ is averaged over all the grid positions such that final score of $D$ is obtained with respect to the input. For testing purposes, we directly use the averaged scores of $\hat{D}$ as a ``detector'' which is run over the grid to detect the abnormality from the input frame (see Sec.~\ref{sec:detection_gan}).

It is important to highlight that both $ \{ F_t \}$ and $\{ O_t \}$ are here collected by using only the frames from{\em normal} scenarios in the identified zones provided by GP. 
The absence of abnormal events at the training phase makes it possible to train the discriminators corresponding to our two tasks without the need of supervised training data: $G$ acts as an implicit supervision for $D$. We hypothesize that the latter lies outside the discriminator's decision boundaries because they represent situations never observed during training and hence treated by $D$ as outliers. We use a \emph{Bank of Discriminators} based on the identified zones provided by GP, which is grouped into two sets: $Set1$, which is trained on a straight path, and $Set2$ that is trained over the curves. The discriminator's learned decision boundaries can be used to detect unseen events as explained in the next section

\subsubsection{Anomaly detection}
\label{sec:detection_gan}

Discriminators are used at the testing phase. More specifically, let $\hat{D}^{F \rightarrow O}$ and $\hat{D}^{O \rightarrow F}$ be the patch-based discriminators trained using the two channel-transformation tasks (see Sec.~\ref{ssec:lgan}). 
Given a test frame $F$ and its corresponding optical-flow image $O$, we first produce the reconstructed $p_O$ and $p_F$ using ${G}^{F \rightarrow O}$ and ${G}^{O \rightarrow F}$, respectively. Then, the pairs of patch-based discriminators  $\hat{D}^{F \rightarrow O}$ and $\hat{D}^{O \rightarrow F}$ are applied respectively to the first and second tasks. Such operation results in two scores for the ground truth observation: $S^O$ and $S^F$, and two scores for the prediction (reconstructed data): $S^{p_O}$ and $S^{p_F}$. The two scores are summed: $S_{observation}= S^O + S^F$, $S_{prediction}= S^{p_O} + S^{p_F}$, and the values in $S_{observation}$ and $S_{prediction}$ are normalized into the range $[0, 1]$. 
%Note that we do not need to produce the reconstruction images to use the discriminators. For instance, for a given position on the grid, $\hat{D}^{F \rightarrow O}$ takes as input a patch $p_F$ on $F$ and a corresponding patch $p_O$ on $O$. 
Note that, a possible abnormality in the observation (e.g., an unusual object/movement) corresponds to an outlier with respect to the data distribution learned by $\hat{D}^{F \rightarrow O}$ and $\hat{D}^{O \rightarrow F}$ during training. The presence of the anomaly results in a low value of $\hat{D}^{F \rightarrow O}(p_F,p_O)$ and $\hat{D}^{O \rightarrow F}(p_O,p_F)$ (prediction), but a high value of $\hat{D}^{F \rightarrow O}(F,O)$ and $\hat{D}^{O \rightarrow F}(O,F)$ (observation). 

Hence, in order to decide whether an observation is abnormal with respect to the scores from the current bank of Discriminators, we simply measure the distance between predicted scores and observation scores such as shown in equation \eqref{eq:dist}.
\begin{equation}\label{eq:dist}
\tilde{Y} = S_{observation} - S_{prediction}
\end{equation}

An error threshold $\tilde{Y}_{thres}$ is defined to detect the abnormal events: when $\tilde{Y}$ is higher than such threshold, the current agent's measurement is considered as an abnormal situation. 

%
%
%\subsection{Ma}
%\label{ssec:m}

\section{Dataset}\label{sec:Dataset}
\begin{figure}[ht] 
	\begin{subfigure}[b]{0.5\linewidth}
		\centering
		\includegraphics[width=3.5cm, height=3.5cm]{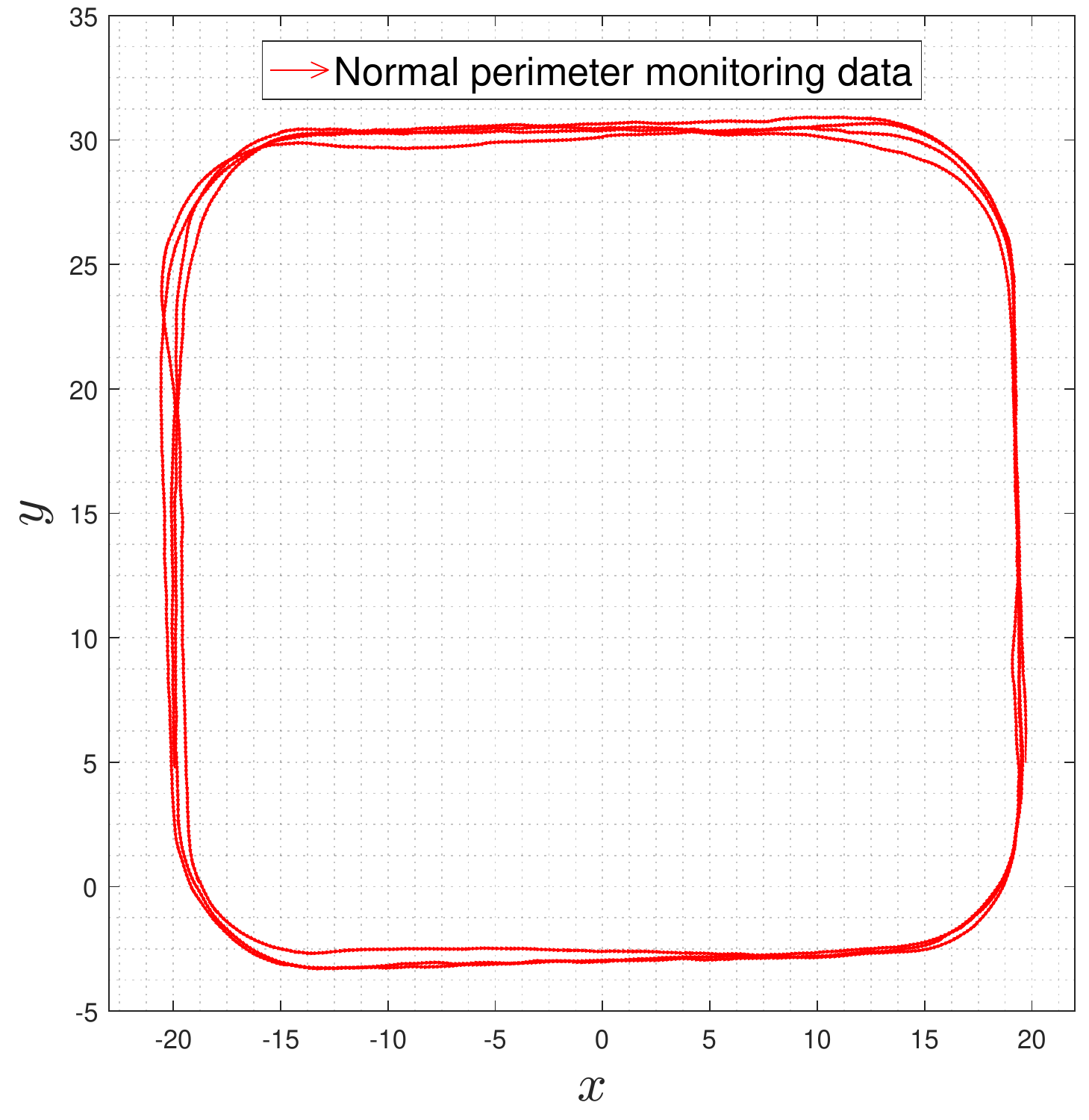} 
		\vspace{-1ex}
		\caption{Perimeter monitoring} 
		\label{fig3:a} 
		%\vspace{2ex}
	\end{subfigure}%% 
	\begin{subfigure}[b]{0.5\linewidth}
		\centering
		\includegraphics[width=3.5cm, height=3.5cm]{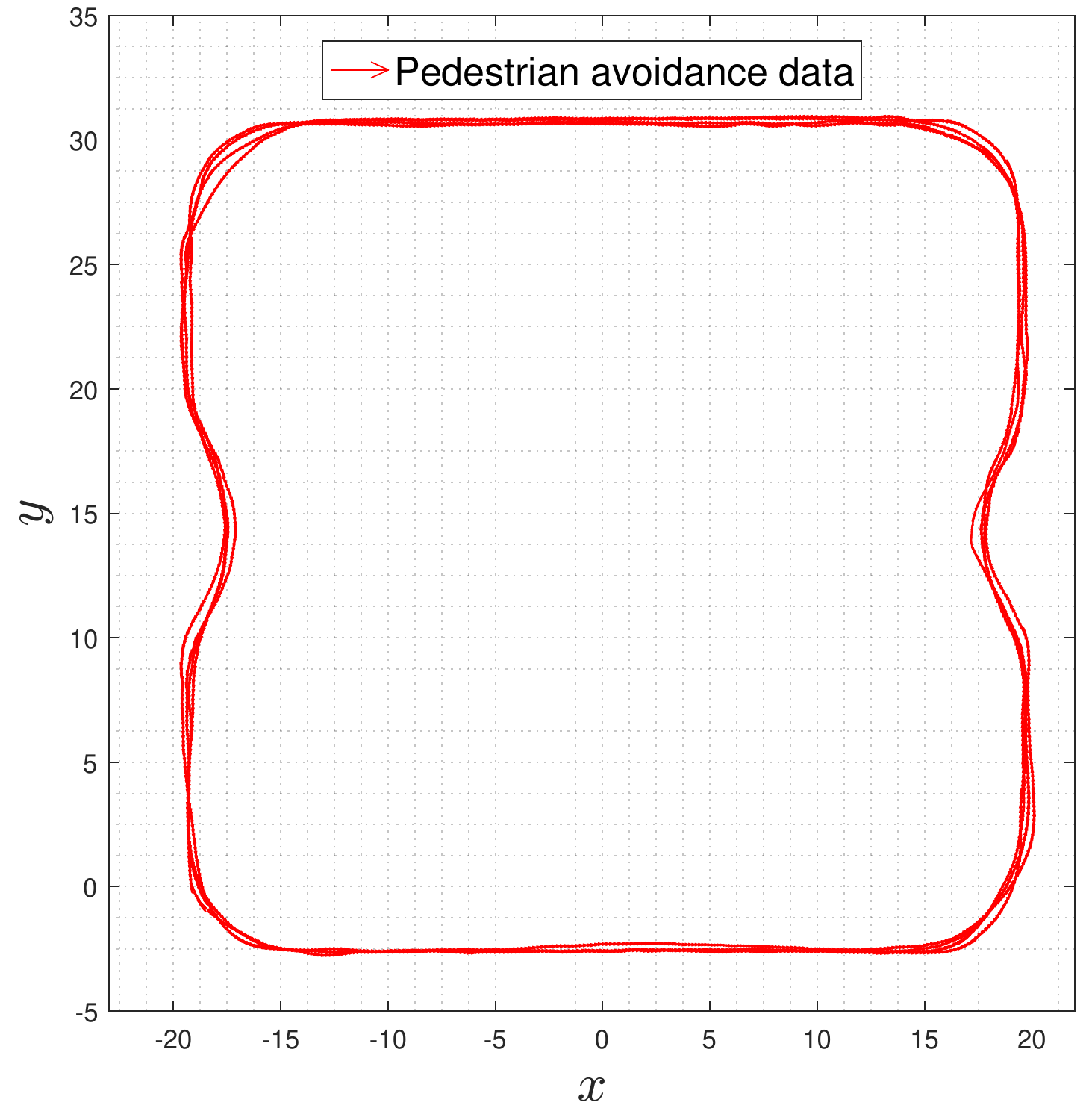} 
		\vspace{-1ex}
		\caption{Pedestrian avoidance} 
		\label{fig3:b} 
		%\vspace{2ex}
	\end{subfigure} 
	  \vspace{-1.5em}
	\caption{Observed tasks in the scene}
	\label{fig3}
	\vspace{-10pt}
\end{figure}

\begin{figure}[ht] 
	\centering
	\includegraphics[width=1\linewidth]{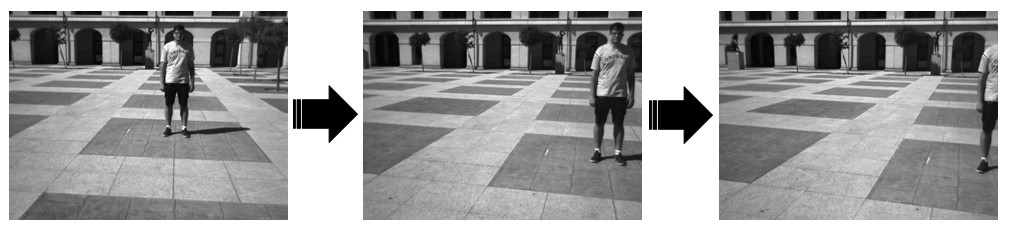} 
	\caption[Odometry]{Pedestrian avoidance from 'iCab' on board camera} 
	\label{fig4} 
	%\vspace{2ex}
\end{figure}

Proposed approach is validated with data acquired from a real vehicle during a perimeter monitoring task. The 'iCab' vehicle is equipped with several heterogeneous sensors \cite{Marin2016}.  In this work, we consider data related to vehicle’s position and images grabbed from a frontal on-board camera
% mapped into Cartesian coordinates provided by the odometry manager, that describes the vehicle's dynamic in the environment. In addition, a binocular camera Bumblebee 2, that provides images of what the vehicle sees.
Two different scenarios are considered, consisting of standard perimeter surveillance task and anomalies while executing it.
\begin{itemize}
	\item Scenario 1: vehicle performs a rectangular path around the environment (perimeter monitoring), see Fig.\ref{fig3:a}.
    \item Scenario 2: vehicle performs an avoidance maneuver to avoid a static pedestrian and then continues the standard patrolling (Fig.\ref{fig4}), see Fig.\ref{fig3:b}.
\end{itemize}
In both scenarios, the vehicle executes the correspondent task several times per each experiment.
%
%\begin{figure}[ht] 
%	\centering
%\includegraphics[width=7cm, height=4cm]{images/icab1.jpg}
%	\caption[Odometry]{Autonomous Driving "iCab"} 
%	\label{fig2} 
%	%\vspace{2ex}
%\end{figure}
%

\begin{figure*}
	\centering
	\includegraphics[width=1\linewidth]{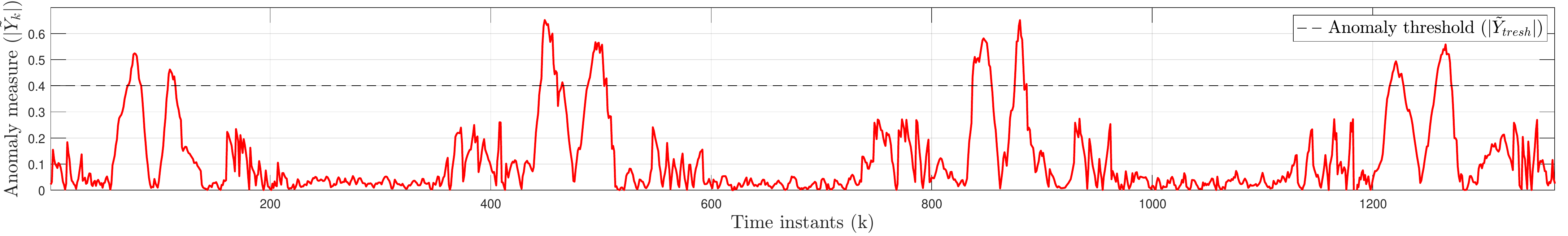}
	\vspace{-1.5em}
	\caption{SL anomaly measurements: perimeter control activity by GP through time with avoidance of static pedestrians.} 
	\label{figAVOID} 
\end{figure*}
\begin{figure*}[t] 
	\centering
	\includegraphics[width=1\linewidth]{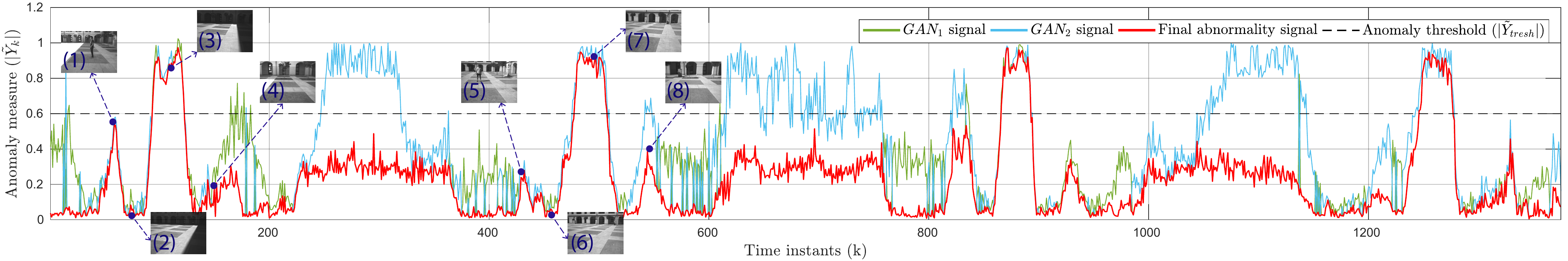}%gan_sig2
	%\vspace{-1.5em}
	\caption{PL anomaly measurements: the distances between the observations and predictions by GANs during the time.} 
	\label{fig:gansig} 
\end{figure*}
\begin{figure}[ht] 
	\centering
	\includegraphics[width=5cm, height=5cm]{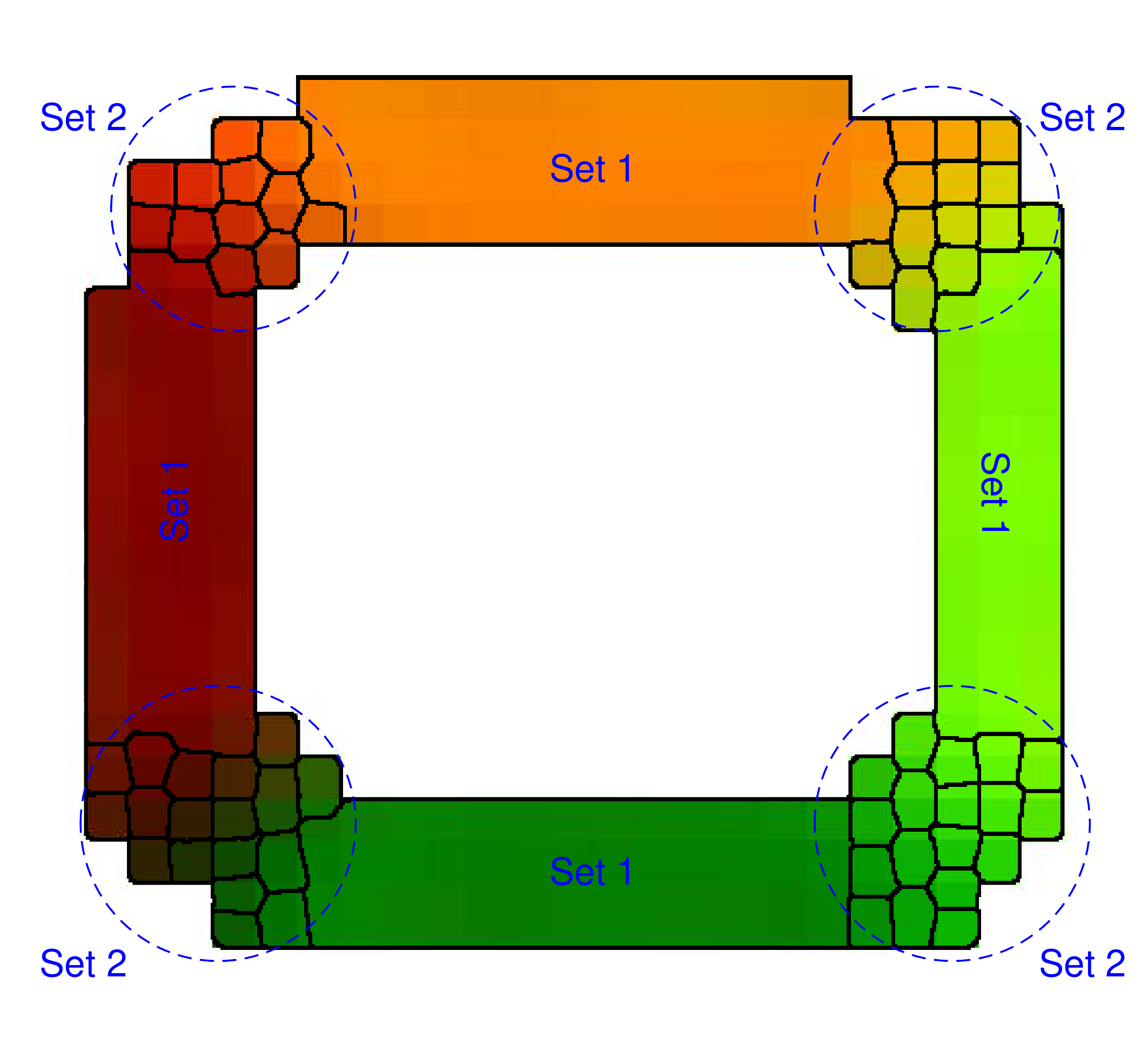} 
	\vspace{-1.5em}
	\caption[Odometry]{Spatial information in terms of zones} 
	\label{Ab_detection} 
\end{figure}
\vspace{-1.5em}
\section{Results}\label{sec:Results}
This section presents the results of proposed methods applied on the dataset presented in section \ref{sec:Dataset}. The normal behavior corresponds to perimeter monitoring defined in the Scenario 1.

\subsection{Shared Level Self Awareness  abnormality detection}
\label{ssec:FMR} 
%a GP estimates the vehicle's velocity components over the scene. the results of GP for $\dot{x}$ and $\dot{y}$  are shown in Fig. \ref{fig3}. Accordingly, we codify GP results into in RGB image that represents the actions over the scene. Additionally, we apply a superpixel approach .
As discussed in subsection \ref{ssec:RDM}, Fig. \ref{Ab_detection} shows the segmentation of GP into zones. In each zone, quasi-constant velocity models are valid. Large and small zones represent the action patterns for going straight and curving respectively. Additionally, each zone is used to create a KF valid in that specific area. As explained in subsection \ref{ssec:AD}, by considering innovations generated by the bank of KFs based on the perimeter control task, it is possible to identify abnormalities simply by observing new trajectory data that does not correspond with the already characterized models. The value of $\xi_k$ measures the abnormality level at the time instant $k$. High innovation values indicate the presence possible anomalies in the scene. By processing position measurements from Scenario 2 (section \ref{sec:Dataset}) and analyzing innovations with respect to the normality model, it is possible to detect anomalies. The abnormality threshold $|\xi_{tresh}|$ is set at $0.4$ and produced anomaly detection results are shown in the Fig. \ref{Ab_detection}. It is possible to find a pattern composed by two abnormal peaks that are associated to the avoidance of the standing pedestrian. An uniform anomaly pattern is not formed due to the straight component of the avoidance maneuver that follows the regular perimeter control behavior. In addition, under the threshold, other two behaviors can be recognized, i.e., straight and curve tracks performed by the vehicle. The lowest abnormality levels correspond to the straight parts of the track. Some abnormality peaks under the threshold value are created when the vehicle curves due to slightly different turning angles when performing the experiment laps.
% Such peaks are explained as, curving to avoid,  going straight and curving To return to the right track, this is why between the two peaks goes under the threshold.
%Two anomaly zones are obtained each
%the peaks under the threshold 

%

\subsection{Private Level Self Awareness abnormality detection}
\label{ssec:SAR}
The bank of GANs are trained on the subsets of data based on GP zones. In our experiments, the bank of GANs is composed of two major subsets: $Set1$ and $Set2$, see Fig. \ref{Ab_detection}. Each GAN detects the abnormality in the corresponding set on which is trained. The self-awareness model is tested on the second scenario discussed in subsection \ref{sec:Dataset}. Anomaly detection results associated to the PL, using the proposed bank of GANs are shown in Fig. \ref{fig:gansig}. Three signals are shown in Fig. \ref{fig:gansig}: The green and blue signals respectively show the computed signals by our $GAN_1$ (trained on $Set1$) and $GAN_2$ (trained on $Set2$). The red signal indicates the final abnormality measurement which is defined as the minimum value of $GAN_1$ and $GAN_2$. As it was expected, the obtained abnormality measurement in PL is aligned with SL results shown in Fig. \ref{figAVOID}.\\
Different parts of the curve can be associated and explained by considering the correspondent images acquired from the on-board sensor. Specifically, the small peak identified with number $1$ can be justified by the presence of the pedestrian in the field of view of the camera: the vehicle do not start the avoidance maneuver yet, thus, it can be seen as a pre-alarm. The small peak in $1$ corresponds to peak in $5$, the latter is smaller due to the posture of the pedestrian, see correspondent images $1$ and $7$.The areas of the curve identified with numbers $2$ and $3$ or $6$ and $7$ correspond to the starting point of the abnormal maneuver and the avoiding behavior itself: it can be seen that peaks $3$ and $7$ are higher than the selected threshold and then correspond to an anomaly. After the small peak $4$, that corresponds to the closing part of the avoidance turn, the vehicle goes back to the standard behavior. In particular, at this point of the curve, the vehicle is actually turning. In the wider area (from $220$ to $380$ secs.), the 'iCab' is moving straight. The slightly higher level of the abnormality curve in straight areas can be explained by a noise related to the vibration of the on-board camera due to the fast movement of the vehicle when increasing its speed.\\
It is notable that, the signal generated by $GAN_1$ becomes higher in the curving areas since it is only trained on $Set1$ for detecting straight paths. Similarly, the $GAN_2$ which is trained on $Set2$, generates higher scores on the straight path. However, both $GAN_1$, and $GAN_2$ can detect the abnormality area (pedestrian avoidance) where both generate a high abnormality score.

\section{Conclusions}\label{sec:Conclusions}
We presented a multi-perspective approach to detect anomalies for moving agents. Obtained results demonstrate the capability of our methodology to recognize anomalies using multiple
viewpoints, namely PL and SL. A future research path could consist in combining information from different sources for decision making and robust the proposed self-awareness model. In particular, situational awareness and self-reactions could be increased with respect to the existing literature.
% Below is an example of how to insert images. Delete the ``\vspace'' line,
% uncomment the preceding line ``\centerline...'' and replace ``imageX.ps''
% with a suitable PostScript file name.
% -------------------------------------------------------------------------

%\vfill
%\pagebreak

%\vfill\pagebreak

%\section{REFERENCES}
%\label{sec:refs}
%
%List and number all bibliographical references at the end of the
%paper. The references can be numbered in alphabetic order or in
%order of appearance in the document. When referring to them in
%the text, type the corresponding reference number in square
%brackets as shown at the end of this sentence \cite{C2}. An
%additional final page (the fifth page, in most cases) is
%allowed, but must contain only references to the prior
%literature.

\bibliographystyle{IEEEbib}
\bibliography{refs} %strings,

\end{document}